\documentclass[nohyperref]{article}

\usepackage{microtype}
\usepackage{graphicx}
\usepackage{subfigure}

\usepackage[dvipsnames]{xcolor}
\usepackage{listings}
\usepackage[inline]{enumitem}

\newcommand\YAMLcolonstyle{\color{red}\mdseries}
\newcommand\YAMLkeystyle{\color{black}\bfseries}
\newcommand\YAMLvaluestyle{\color{blue}\mdseries}

\makeatletter

\newcommand\language@yaml{yaml}

\expandafter\expandafter\expandafter\lstdefinelanguage
\expandafter{\language@yaml}
{
  keywords={true,false,null},
  keywordstyle=\color{darkgray}\bfseries,
  basicstyle=\YAMLkeystyle,                                 
  sensitive=false,
  comment=[l]{\#},
  morecomment=[s]{/*}{*/},
  commentstyle=\color{purple}\ttfamily,
  stringstyle=\YAMLvaluestyle\ttfamily,
  moredelim=[l][\color{orange}]{\&},
  moredelim=[l][\color{magenta}]{*},
  moredelim=**[il][\YAMLcolonstyle{:}\YAMLvaluestyle]{:},   
  morestring=[b]',
  morestring=[b]",
  literate =    {---}{{\ProcessThreeDashes}}3
                {>}{{\textcolor{red}}}1     
                {|}{{\textcolor{red}}}1 
                {\ -\ }{{\mdseries\ -\ }}3,
}

\lst@AddToHook{EveryLine}{\ifx\lst@language\language@yaml\YAMLkeystyle\fi}
\makeatother
\usepackage{booktabs} 

\usepackage{hyperref}

\usepackage[accepted]{icml2022}

\usepackage{amsmath}
\usepackage{amssymb}
\usepackage{mathtools}
\usepackage{amsthm}

\usepackage[capitalize,noabbrev]{cleveref}

\theoremstyle{plain}

\theoremstyle{definition}

\theoremstyle{remark}

\usepackage[textsize=tiny]{todonotes}

\begin{document}

\twocolumn[
\icmltitle{Synthetic Dataset Generation for Adversarial Machine Learning Research}



\icmlsetsymbol{equal}{*}

\begin{icmlauthorlist}
\icmlauthor{Xiruo Liu}{equal,yyy}
\icmlauthor{Shibani Singh}{equal,yyy}
\icmlauthor{Cory Cornelius}{yyy}
\icmlauthor{Colin Busho}{comp}
\icmlauthor{Mike Tan}{comp}
\icmlauthor{Anindya Paul}{yyy}
\icmlauthor{Jason Martin}{yyy}
\end{icmlauthorlist}

\icmlaffiliation{yyy}{Intel Corporation, Hillsboro, OR, USA}
\icmlaffiliation{comp}{The MITRE Corporation, McLean, VA, USA}

\icmlcorrespondingauthor{Xiruo Liu}{xiruo.liu@intel.com}

\icmlkeywords{Machine Learning, ICML}

\vskip 0.3in
] \printAffiliationsAndNotice{}



\begin{abstract}
Existing adversarial example research focuses on digitally inserted perturbations on top of existing natural image datasets.
This construction of adversarial examples is not realistic because it may be difficult, or even impossible, for an attacker to deploy such an attack in the real-world due to sensing and environmental effects.
To better understand adversarial examples against cyber-physical systems, we propose approximating the real-world through simulation.
In this paper we describe our synthetic dataset generation tool that enables scalable collection of such a synthetic dataset with realistic adversarial examples.
We use the CARLA simulator to collect such a dataset and demonstrate simulated attacks that undergo the same environmental transforms and processing as real-world images.
Our tools have been used to collect datasets to help evaluate the efficacy of adversarial examples, and can be found at \url{https://github.com/carla-simulator/carla/pull/4992}.
\end{abstract}

\section{Introduction}
\label{sec:intro}
Deep Neural Networks (DNN) have revolutionized the field of artificial intelligence with significant success in many emerging fields like computer vision and natural language processing.
As the production and deployment of DNN models are on the rise in security and safety critical applications, numerous studies have shown that DNN models are susceptible to adversarial examples \cite{Szegedy2014,Carlini2017,Goodfellow2015,Papernot2016}. 

\begin{figure}[!ht]
    \subfigure[Green Patch in CARLA]{\label{fig:patch_green}\includegraphics[width=0.23\textwidth]{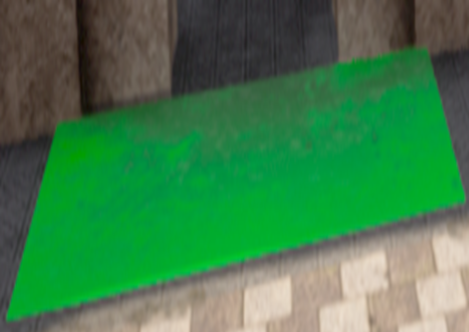}}
    \hfill
    \subfigure[Digital Patch]{\label{fig:patch_digital}\includegraphics[width=0.23\textwidth]{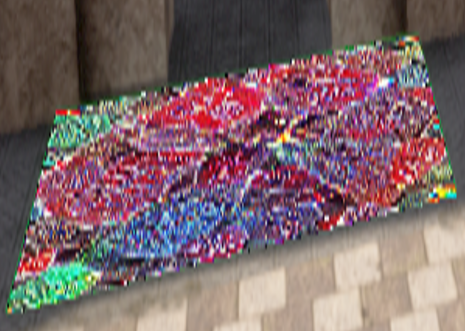}}
    \hfill
    \\
    \subfigure[DAPRICOT Patch]{\label{fig:patch_dapricot}\includegraphics[width=0.23\textwidth]{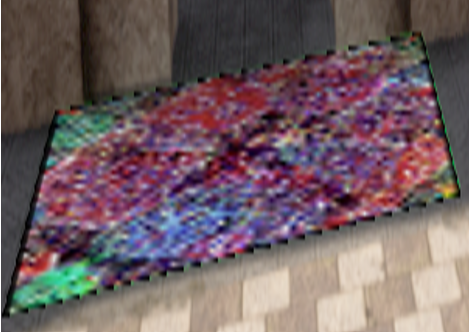}}
    \hfill
    \subfigure[Rendered Adversarial Patch]{\label{fig:patch_carla}\includegraphics[width=0.23\textwidth]{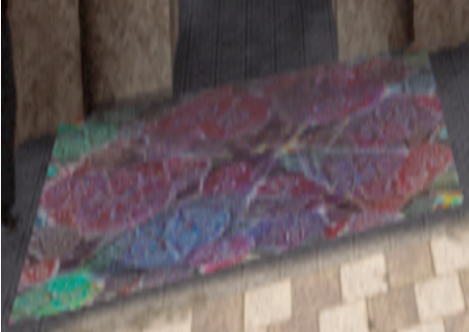}}
    \caption{A comparison of patch insertion methods. \subref{fig:patch_digital} shows a digital adversarial patch that is composed directly onto the image and obviously does not blend into the surrounding environment. In \subref{fig:patch_dapricot}, a DAPRICOT \cite{9762099} patch is crafted to mimic environmental effects, but artifacts are still noticeable. In \subref{fig:patch_green}, a green patch is first inserted into CARLA, then an adversarial texture is streamed onto it as shown in \subref{fig:patch_carla}. Hence, this patch is rendered by CARLA in the same way as other objects in the scene.}
  \label{fig:ZoomedPatches}
\end{figure}

In the computer vision domain, although the robustness of defense schemes are often analyzed through digital perturbation attacks, these attacks are very difficult to deploy in practice as they manipulate images directly after the inputs have been captured and digitized inside the system.
Digital insertion attacks are not realistic as they do not account for real-world environmental transforms, such as shading, lighting, occlusions and sensor noise.
\Cref{fig:ZoomedPatches} shows an exemplar comparison between the realism of three types of patch attacks.
Other work has shown that physical adversarial examples need to be robust against multi-distance and multi-perspective view of the camera in order to have an impact in a real-world scenario  \cite{Lu2017NONT}.
Hence, it is not a fair evaluation for many defense schemes, if any at all, with these types of digital attacks.

We propose a simulation tool that is able to generate large scale synthetic datasets with realistic adversarial examples.
By leveraging the capabilities of CARLA \cite{Dosovitskiy17}, our simulated datasets have the following distinct attributes: multi-modal sensory measurements, instance segmentation labels, diverse backgrounds with varying weather conditions, and multi-perspective views of the scenes.

\section{Related Work}
\label{sec:related_work}
There has been a significant amount of literature on generating synthetic datasets for various applications, including autonomous driving through diverse scenes with synthetic pedestrians and objects \cite{geiger2012we,Cordts2016Cityscapes,Richter2016PlayingFD}.
For example, Sim~200k \cite{Johnson-Roberson2017} dataset consists of $200$k synthetic images of vehicles driving in different times of the day under diverse weather and lighting conditions.
A lot of recent work relies on synthetic data to create benchmarks, or to compensate for the lack of training data with data augmentation.
SynDataGeneration \cite{dwibedi2017cut} cut and paste object patches in scenes to create images thus focusing on patch-level realism. This technique requires preexisting object masks  which are generated using pixel-objectness models with bi-linear pooling. 

Prior work \cite{46561} have shown that it is possible to generate scene-independent adversarial patches which can be placed anywhere and is able to alter the behaviour of the targeted model.
Generating patches, which work under various lighting conditions and transfer to different target models, is a computationally intensive optimization process.
More recent work \cite{Pintor2022ImageNetPatchAD} released a small adversarial patch dataset for fast robustness evaluation, where patches are applied with translation and rotation and composed onto images from ImageNet.
Imagenet-C \cite{hendrycks2019benchmarking} and WILDS \cite{koh2021wilds} datasets address distribution shifts in the dataset.
Imagenet-C has image corruptions (e.g., blurring, jitter) and WILDS addresses domain generalization and sub-population shifts.

APRICOT open source dataset \cite{braunegg2020apricot} was the first step to provide a benchmark to evaluate the robustness of object detection models against realistic attacks.
APRICOT photographed printed adversarial patches on real objects and scenes in real-world environments. However, there are several drawbacks of this dataset: not every object in the image is labelled; there is no complexity to the attacks since the attacks are static and not adaptive in nature; not all patches work against all models. Moreover, this manual data collection is time-consuming, not configurable to new environmental factors and does not capture multi-modal inputs beside RGB images.
Dynamic APRICOT (DAPRICOT) \cite{9762099} is one step further towards a realistic adversarial dataset using CARLA with customized assets.
Customized green screen patches, which serve as the place holders for adversarial patches, are inserted into the scene.
Then, color transforms these patches undergo are taken and used for adjusting  adversarial patches such that the adversarial patches blend into the simulated environment more realistically.
While a step closer to realistic attacks leveraging simulation, this attack cannot map textures to 3D objects.
Prior work \cite{Cornelius2019TalkPT} showed how to use CARLA to demonstrate the efficacy of ShapeShifter \cite{chen2018shapeshifter} attacks against driving scenarios.
That work also highlighted the importance of validating realistic attacks by reproducing scenarios across varying environmental conditions, which CARLA already supports.

There are other advancements of placing adversarial attacks physically and capturing images in different weather conditions.
For example, with the sign embedding attack \cite{sitawarin2018rogue}, traffic signs are modified to improve robustness to noisy transformations happened during the image capture stage.
This attack also exemplifies the difficulty in creating datasets with realistic attacks.

Privacy concerns also raise for real-world datasets collected from public places, especially for tasks like pedestrian detection and tracking. 
To address privacy concerns, generating a large diverse synthetic dataset using a rendering game engine is explored in \cite{fabbri2021motsynth}.
The dataset has temporally consistent bounding boxes, instance segmentation, depth maps and pose occlusion information along with varied environment conditions, camera viewpoints, object textures, lighting conditions, weather, seasonal changes, and object identities.

\section{Motivation}
\label{sec:motivations}

While aforementioned datasets maybe useful for the purposes of augmenting benign datasets, there is an imminent need for frameworks that enable the exploration of realistic adversarial attacks with varying threat models.

\begin{figure}[th]
    \centering
    \includegraphics[width=1.0\linewidth]{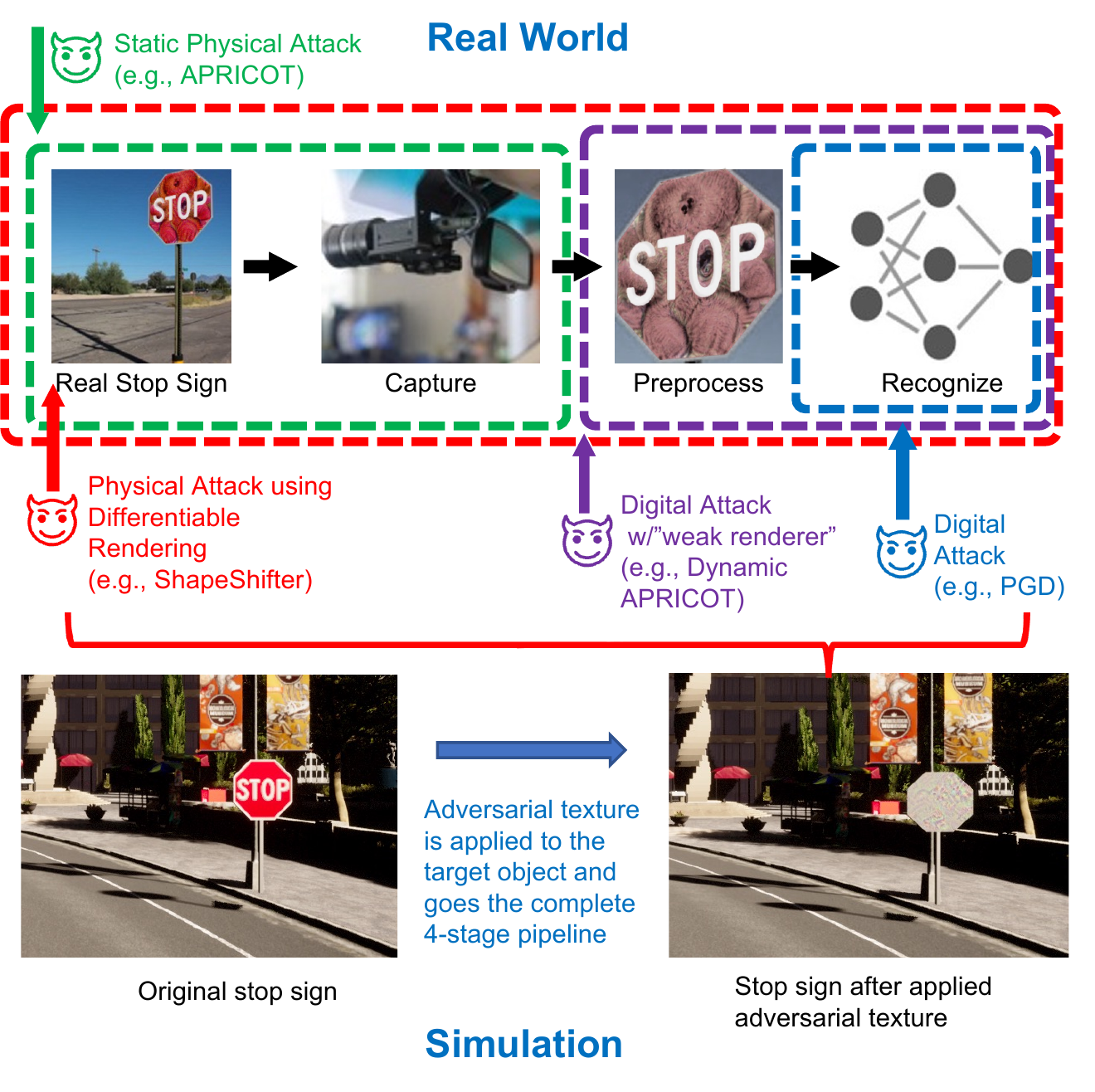}
    \caption{Comparison of performing attacks in the real world versus in the simulation. The end-to-end image classification pipeline includes four stages. As shown at the top, in the real world, a realistic attack needs to be applied to a real object to undergo the complete pipeline, which makes it very challenging. However, with simulation, it is convenient to apply a adversarial texture to the target stop sign and force the attack to undergo all the same environmental transformations and processing as the defense model.}
  \label{fig:attack_pipeline}
\end{figure}

We define \textit{realistic attacks} as those attacks that use the same end-to-end processing pipeline as the defense models.
In \Cref{fig:attack_pipeline}, the complete pipeline for image classification tasks typically includes four stages.
Attacks may come in at different stages.
For example, digital attacks such as Projected Gradient Descent (PGD)~\cite{madry2017towards} are injected at the final recognition stage, while DAPRICOT attacks are performed at preprocessing stage.
Physical attacks and static physical attacks are injected at the very front and applied to real-world objects, which makes them realistic.
However, these physical attacks are very expensive and hence are costly to scale.
Therefore, we adopt an alternative approach by replacing the real-world with a simulated world created by CARLA.
As a result, we can apply adversarial textures onto 3D objects, which then undergo all environmental transforms in the simulation as shown in \Cref{fig:attack_pipeline}.

\section{Data Collection Tool}
\label{sec:tool}

Motivated by the lack of large scale dataset with realistic adversarial examples, we developed a scalable data collection tool that is built on top of CARLA.
We open sourced this tool at \url{https://github.com/carla-simulator/carla/pull/4992}.

\begin{figure}[th]
    \centering
    \includegraphics[width=1.0\linewidth]{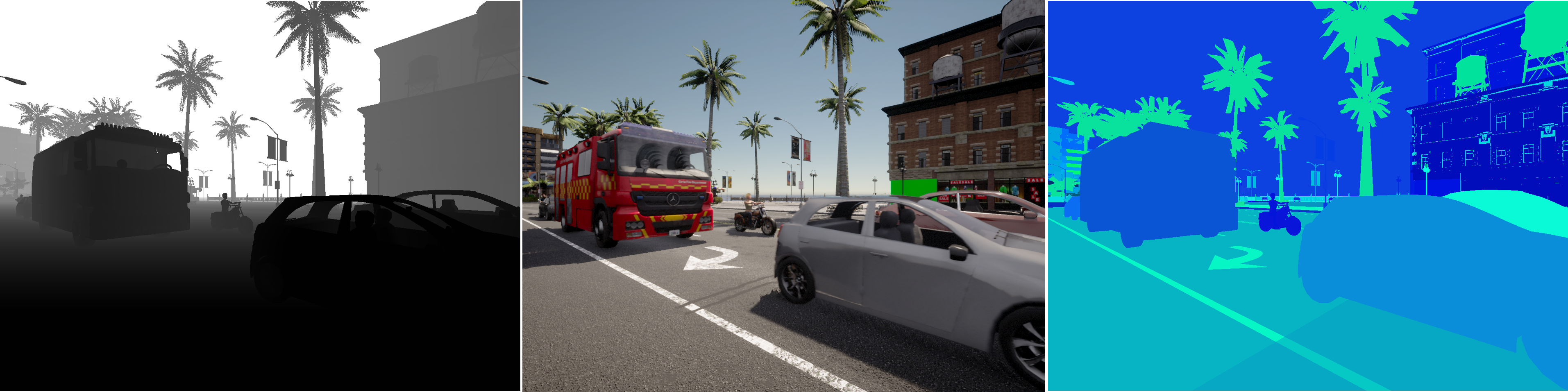}
    \caption{Multimodality is well supported in CARLA, while it may be very difficult to obtain synced data from different sensors in a real-world dataset. From left to right, the depth image, the RGB image and the ground truth (i.e., instance segmentation masks) provide well aligned data for research that explores the relations among different modalities. }
  \label{fig:multimodality}
\end{figure}

This tool enables the collection of multimodal data (as shown in \Cref{fig:multimodality}) and different scenes from sensors placed at different locations.
A wide range of scenarios can be configured to collect data with varying weather patterns, times of day, traffic, and crowd conditions. The tool provides accurate labels and minimizes the costs involved in generating annotations for the dataset. The annotations include bounding boxes, classification labels, instance segmentation masks and frame sequence order for temporal tasks.

CARLA supports various atmospheric conditions and illumination settings with tunable configurations for position and color of the sun, the intensity and color of diffuse sky radiation, ambient occlusion, atmospheric fog, cloudiness, precipitation, as well as lighting conditions by time of day. CARLA allows configurable placements of sensors from different modalities, such as RGB, depth, LIDAR, RADAR, optical flow, event based dynamic vision sensor and ground-truth sensors (instance segmentation and semantic segmentation). Taking advantage of these features from CARLA, we developed a data collection tool, which supports various configuration of synchronized sensor suites, weather and lighting conditions, traffic and crowd settings. 

The tool consists of two major components: a data saver and an annotator. The data saver takes a configuration file (details in \Cref{lst:config_yaml}) as an input and then instantiates a CARLA simulation accordingly. We support three kinds of sensor placements: statically placed sensors, sensors attached to vehicles or pedestrians, and sensors that move.

The three supported sensor movement patterns are:
\begin{enumerate*}[label=(\arabic*)]
    \item linear motion, where the sensor moves straight from the source position to the destination position;
    \item rotation, where the sensor rotates within a predefined angle range;
    \item jitter, where the sensor jitters randomly within a configured range.
\end{enumerate*}
All the three types of movement can be applied together to a moving sensor. Also, jitter and rotation can be added to a ``static'' sensor too. 

\begin{figure}[ht]
    \subfigure[Scene]{\label{fig:scene}\includegraphics[width=0.48\linewidth]{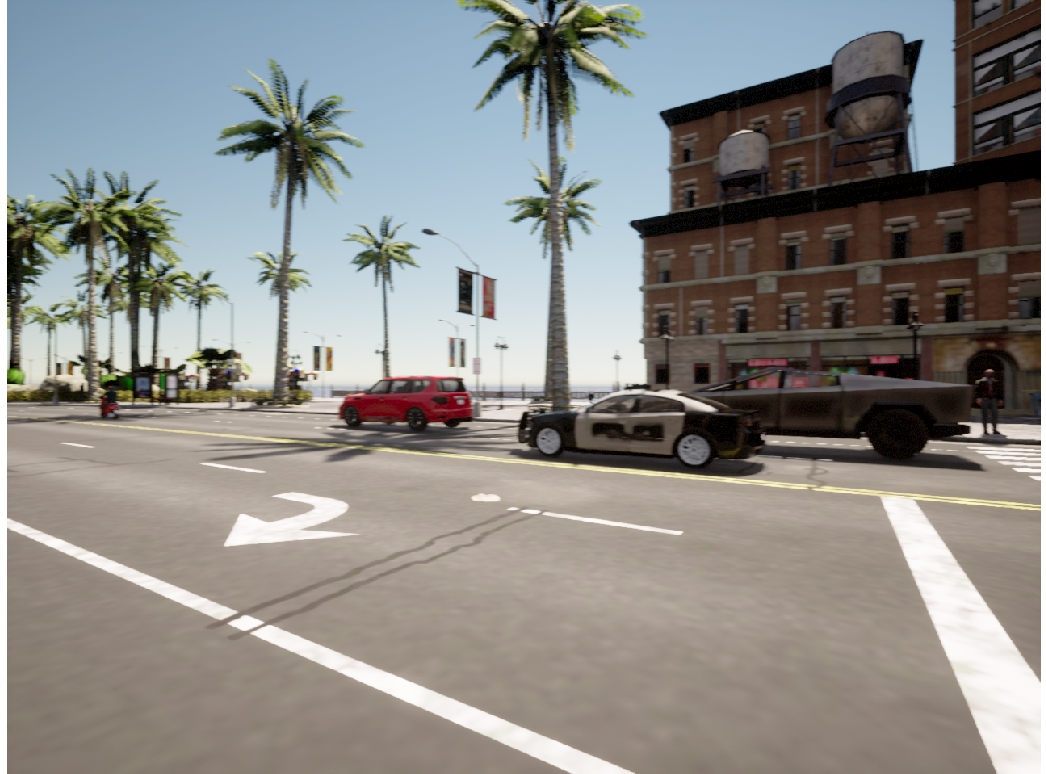}}
    \hfill
    \subfigure[Annotated scene]{\label{fig:annotated_scene}\includegraphics[width=0.48\linewidth]{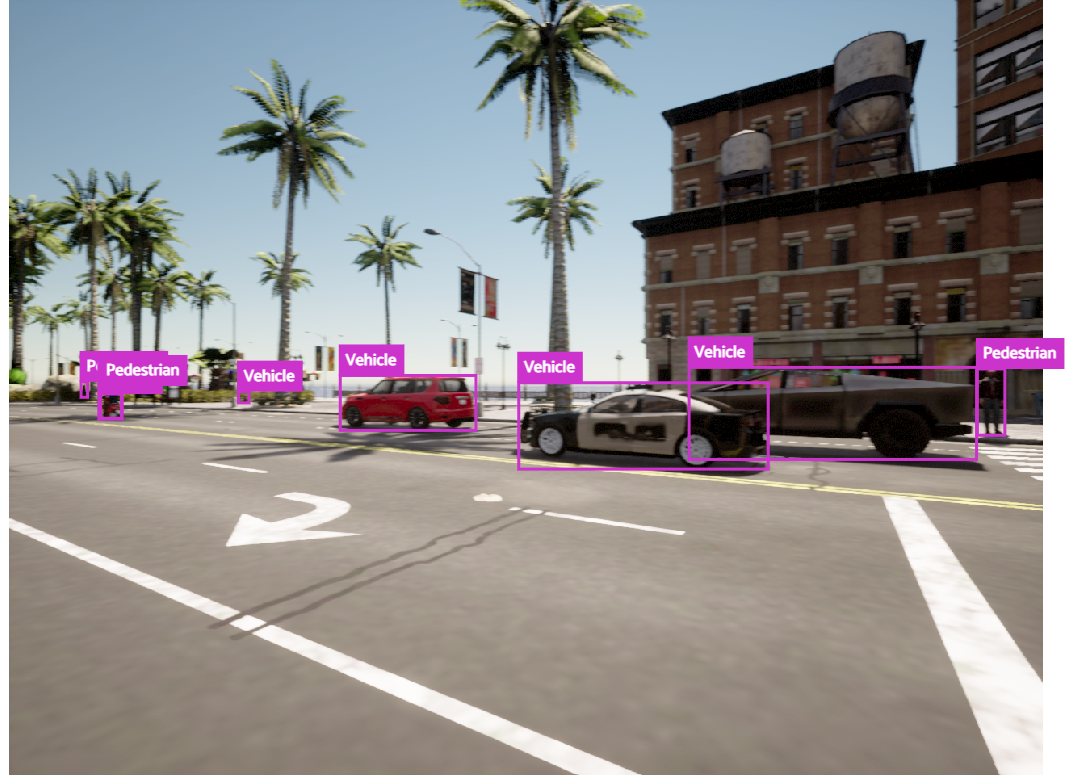}}
    \caption{An image captured by a RGB camera from CARLA using the data saver and the visualization of the generated annotations using the annotator tool: \subref{fig:scene}~scene and \subref{fig:annotated_scene}~annotated scene.}
  \label{fig:annotated_image}
\end{figure}

The annotator supports two types of annotation formats: kwcoco \cite{kitware2020kwcoco} and Multi Object Tracking and Segmentation (MOTS)~\cite{voigtlaender2019mots}. The kwcoco annotation format is an extension of the COCO format with COCOAPI support. We choose this format to support temporal annotations that are not supported in COCO. The MOTS format has two forms, text and PNG, to support object tracking and segmentation scenarios. \Cref{fig:annotated_image} shows a RGB image and the visualization of its annotations generated by our tool. 

We focus on five specific aspects that we find lacking in the existing datasets and synthetic data generation frameworks: configurability with different settings via a YAML configuration file, reproducibility of data with a seed and configuration, the enablement of realistic attacks that go through the same environmental transforms as the defenses, extensibility to augment a existing dataset to accommodate new requirements, ease of use with docker to bypass the complexity from CARLA and Unreal Engine installations and simplify running simulations, and scalability to be able to run multiple simulations in containers simultaneously on different GPUs.

\section{Capabilities to Enable Benign and Adversarial Machine Learning Research}
\label{sec:capabilities}

Configurability and extensibility have been of huge importance to the development of datasets that advance machine learning research. Quite often, when attempting to deep dive into complex problems and decouple a set of mingled factors, researchers find out that it is very challenging to obtain new customized data from the real world that can help them conduct experiments and pinpoint the root cause. They may have to use the initial real world dataset due to the heavy cost in collecting and labelling that specific data. 

However, with our tool, extending an existing dataset with customized configurations to meet new research requirements become feasible. Because the simulation can isolate one environmental factor at a time, researchers can determine specific effects that lighting, weather, object orientation, or time of day have on a model's ability to correctly perform its task by easily creating new data that only changes a chosen environmental factor. As a result, researchers can have better understanding of the model and hence improve the model correspondingly. Moreover, as the data collection is defined by configurations in our tool, it is easy to extract the statistics of the dataset so that we can better understand the characteristics and limitations of the models trained on the dataset. And if more data is needed at a later time, it is also convenient to generate more data with new customized configurations.  

\begin{figure}[th]
    \centering
    \includegraphics[width=1.0\linewidth]{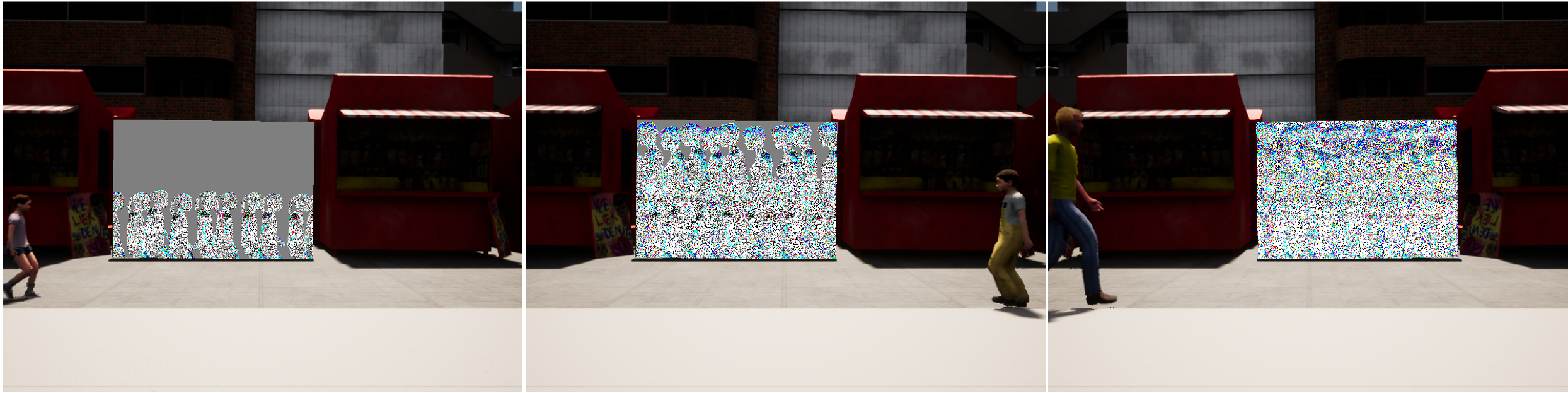}
    \caption{An example of how the simulation tool helps a patch attack generalizes better with more object tracking data. The left figure shows that the attack overfits to a single pedestrian tracking video. In the middle figure, the attack improves as it is computed over five tracking videos. The attack shown in the right figure is optimized with ten diverse tracking videos and hence performs the best among the three.}
  \label{fig:attack_generalization}
\end{figure}

From an adversarial perspective, an attacker can also enjoy all the benefits. Just as model training may overfit to a given training dataset, an attack may overfit to a set of data as well. By being able to easily create additional data and change environmental factors, an attacker can generalize the attack better to a given situation. \Cref{fig:attack_generalization} shows an example of how a patch attack may overfit to an object tracking model in a scenario, where a child walks by a wall with an adversarial patch on it. As more diverse data is created from simulations, however, the attack improves and generalizes better as it is able to attack the tracking model in case of any person walking in front of the adversarial patch on the wall from any directions. 

\begin{figure}[th]
    \centering
    \begin{subfigure}
        \centering
        \includegraphics[width=1.0\linewidth]{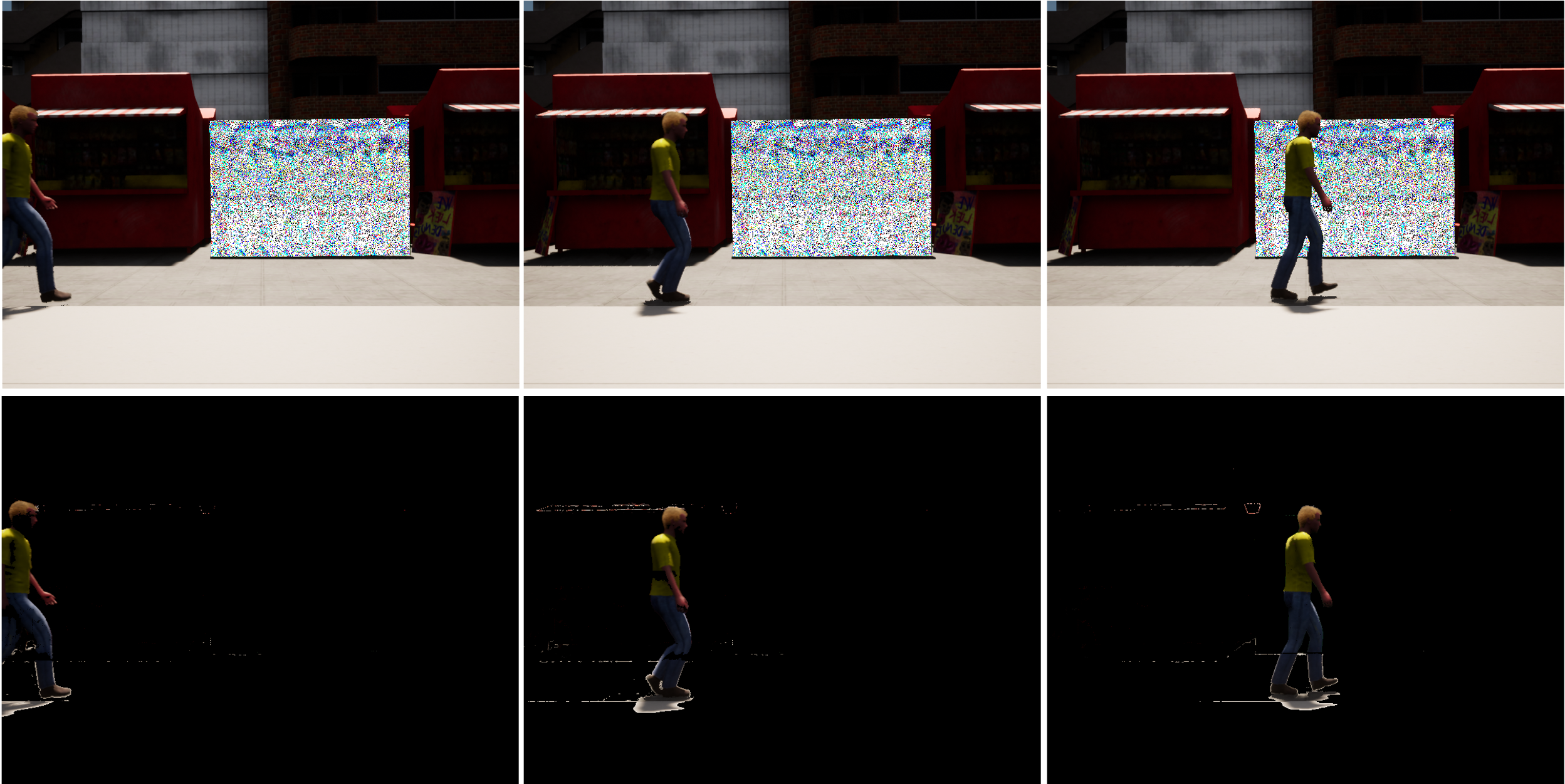}
    \end{subfigure}
    \begin{subfigure}
        \centering
        \includegraphics[width=1.0\linewidth]{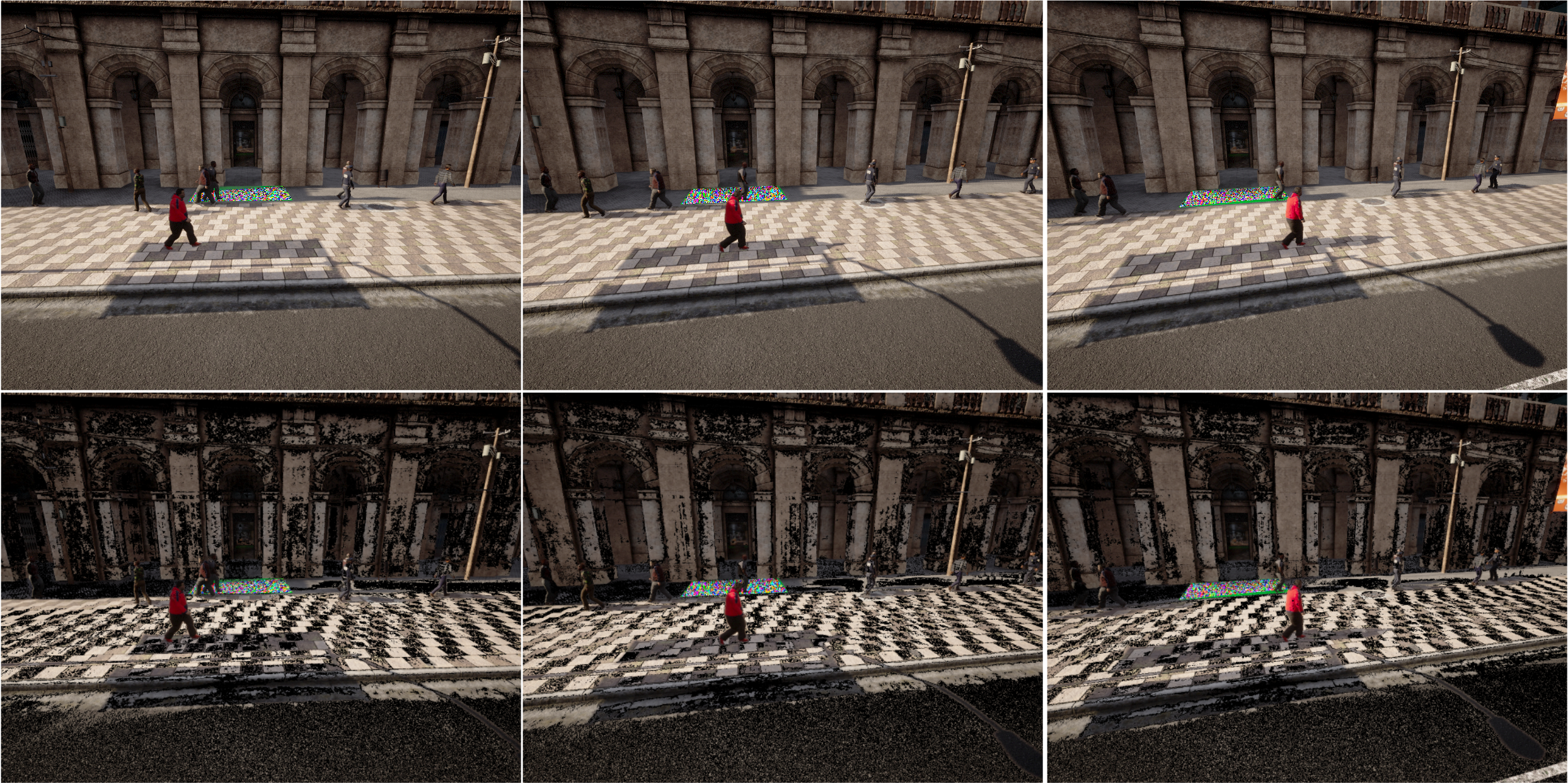}
    \end{subfigure}
    \caption{Generalization of the adversarial patch defense using background ablation. Figures at top two rows show that an adversarial attack on a video with a static background, and the defense successful ablates the adversarial patch. However, as shown by the figures in the bottom two rows, an attack on a tracking video with dynamic background causes the defense fail to ablate the adversarial patch.}
  \label{fig:background_ablation}
\end{figure}

Researchers can test and challenge assumptions for machine learning defenses and attacks, when given the ability to create synthetic data in this manner. As an example, \Cref{fig:background_ablation} shows a machine learning defense that assumes a static background. The figures in the first row show the scene where an adversarial patch is placed on the background wall. The figures in the second row show the defense mechanism of ablating the background in order to remove the patch. In the static setting, it is clear that this defense scheme is quite effective by masking out the background. However, a more advanced attack may challenge this static assumption and test how well the defense holds up under a different assumption. The third row of \Cref{fig:background_ablation} shows a set of images, which is a moving scene as the camera changes its transform slightly. The figures in the fourth row show how the defense performs in this dynamic setting. The adversarial patch is not removed and the defense model fails to ablate the background as expected. 
This experiment shows that even a slight shift in the background of the scene causes the background ablation defense to be largely ineffective. This type of study would be time consuming and expensive, if at all possible, in a real world setting. A researcher would need to recollect a scene, ideally keeping most of the environmental factors as similar as possible. Using the data collection tool we introduced allows the recollection of a scene  with a simple change (e.g. adding sensor jitter movement) to a configuration file. This allows researchers to better understand what makes their machine learning defenses generalize better.

\begin{figure}[!ht]
    \centering
    \includegraphics[width=1.0\linewidth]{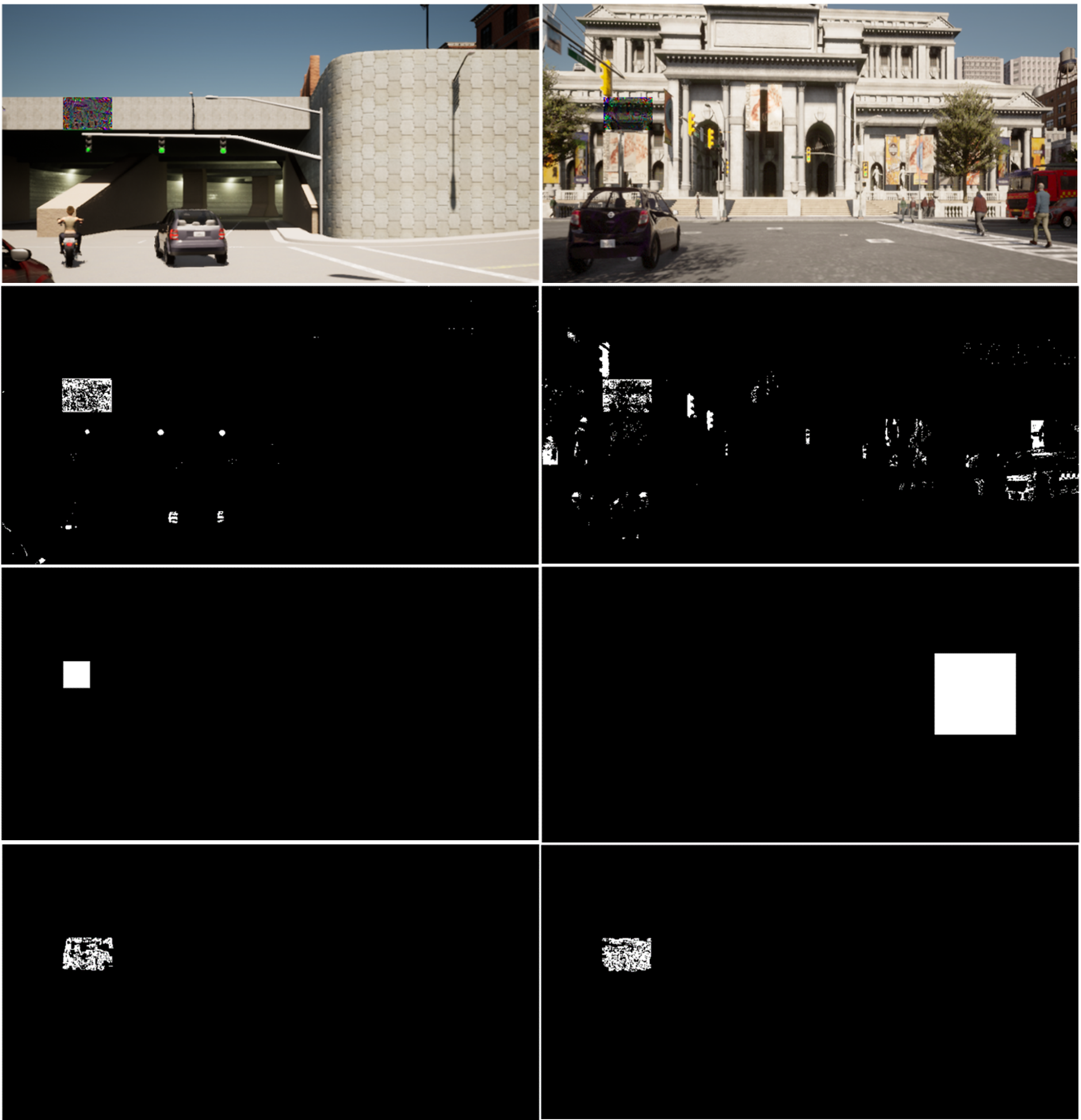}
    \caption{Generalization of adversarial patch defense using physical constraints \cite{feng2022constraining}. The left column includes a low resolution RGB image and the visualization after applying the defense, while the right column is a high resolution RGB image with the visualized defenses. The first row shows two adversarial images with an adversarial patch in the top left quadrant. The second row shows masks of pixels with anomalous colors. The third row shows masks in regions with high frequency contents. And the fourth row shows masks in regions with high hue and saturation pixel values. This defense, which was trained on low resolution images, performs well on the left column images as it is able to identify the patch location. However, it does not generalize well on high resolution images as shown in the right column where the patch is located at wrong position or false positives occur.}
  \label{fig:physical_constraints_generalization}
\end{figure}

As another example, \Cref{fig:physical_constraints_generalization} demonstrates how an attack's efficacy may vary when placed in different scenes by taking into consideration of physical constraints in the scene. An adversarial patch defense may attempt to locate a patch by detecting anomalous natural scene statistics. The success of the defense largely depends on where other objects are located within the scene as well as the natural environmental statistics of the scene. \Cref{fig:physical_constraints_generalization} provides a comparison on how a patch defense performs on low resolution images and high resolution images. As it was trained on low resolution images, the defense performs well on the RGB image in the left column and can locate the patch as shown in the other three images in the left column. However, given a high resolution RGB image in the right column, the defense performance degrades as the patch is located at a wrong place in the third image in the right column. While the defense still identifies the patch in the second image, it also generates many false positives as well. These failures indicate potential generalizability issues of this defense that uses physical constraints.

Our tool can help better generalize defenses by enabling researchers to easily generate additional data with variability for different scenarios. On the other hand, attackers could also receive benefits as they can take this information and use it to their advantage. For example, an adversarial patch can be placed in a location that better ``blends in'' to the scene. Or the patch may be optimized to more closely match the current environment where it is placed.    

Multimodal machine learning is one of the most vibrant research areas that aims to process and learn related information from different modalities. As it is similar to how humans sense and learn the world, multimodal learning shows increasing importance and extraordinary potential. However, a big challenge in this research area is the lack of high quality, well aligned data from different modalities. It is very difficult to obtain real-world datasets with data captured from different types of sensors at the exact same location and perspective. On the other side, it is easy to generate a multimodal synthetic dataset using CARLA, which currently supports RGB camera, depth camera, optical flow camera, LIDAR and RADAR sensors. \Cref{fig:multimodality} shows an example where a depth camera and a RGB camera are perfectly aligned (i.e., placed at the exact same location with the same rotation) and share the same ground truth.  

In the adversarial machine learning domain, digital perturbation attacks, such as PGD, are widely used for evaluating the robustness of models. However, these attacks are often too powerful in the sense that they directly attack the image pixels after the image has been captured and hence do not undergo the environmental transforms and the processing pipeline that a defense scheme needs to consider for accuracy and robustness. 

We show a comparison between varying methods of inserting an adversarial patch into a scene captured by CARLA in \Cref{fig:ZoomedPatches}. The digital patch in \Cref{fig:patch_digital} is composed directly onto the image, which bypasses all the environmental transforms. Note that there are no shadows on the patch even though it is located in the shadowy area of the sidewalk. \Cref{fig:patch_dapricot}  shows a DAPRICOT patch that estimates the environmental transforms and calculates a color correction to be applied to the patch before digitally inserting into the scene. Finally, using CARLA, a green patch in \Cref{fig:patch_green} is first inserted into the scene as a place holder. Then the adversarial patch is applied as a texture to the designated area in the simulation, and then rendered in CARLA as shown in \Cref{fig:patch_carla}. This approach makes the patch look very realistic as it takes the environmental transforms (such as shadow, lighting, reflections) and applies them to the patch in the same way as other objects in the scene. To evaluate the robustness of machine learning models fairly and accurately, it is critical to perform realistic attacks that undergo the complete processing pipeline as normal models would take. Our synthetic dataset generation has the capability to force the attacker to consider the effects of environmental transformations. As a result, we are able to provide an even playground for both the attacker and the defender.  

\begin{figure}[th]
    \centering
    \includegraphics[width=1.0\linewidth]{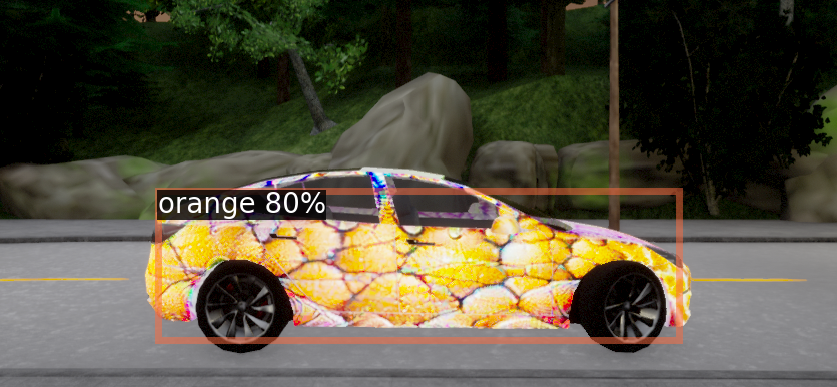}
    \caption{Validation of a ShapeShifter attack in CARLA. An adversarial texture is computed and then applied to a Tesla Model~3 car in the simulation. This attack is successful as it causes the vehicle to be misclassified as an orange.}
  \label{fig:orange_car}
\end{figure}

Moreover, from the attacker's perspective, in real life it is very expensive and slow to compute and apply a physical adversarial perturbation. However, with the simulation tool, it is much faster to iterate the perturbation computations and evaluate the effectiveness of the realistic attack. The tool also allows an attacker to choose any suitable technique to craft adversarial examples for specific scenes.  \Cref{fig:orange_car} shows an example of validating a ShapeShifter \cite{chen2018shapeshifter} attack, where an adversarial texture is computed and then applied to a Tesla Model~3 car in CARLA. This attack is successful as it causes the vehicle to be misclassified as an orange. Without simulation, to validate this physically realizable attack, a car needs to be painted with the computed adversarial texture to validate the effectiveness of the attack in the real world. Even worse, the attack optimization may go through this \textit{paint-validation-repaint} process multiple times, which makes it very expensive to conduct realistic attack research in real world. On contrary, our simulation tool brings significant benefits for developing realistic attack techniques, as it saves the heavy overhead of creating real adversarial objects while at the same time ensures the attack undergoes the same environmental transformations as the defense.  

Hence, our tool provides a cheap and convenient alternative to the real-world adversarial dataset generation, which has an prohibiting cost for realistic adversarial examples and thus is very difficult to be of large scale, if even possible. With the simulation tool, it becomes much easier to insert 2D or 3D adversarial perturbations back into the simulation. And due to its low cost and low time consumption, generating a large scale synthetic adversarial dataset with realistic attack samples becomes feasible. Even further, as discussed in \cref{fig:ZoomedPatches}, streaming texture capability provided by CARLA, which allows the texture of an object to be changed in real time during the simulation, opens an door for the adversarial training with realistic adversarial noises.

\section{Conclusion}
\label{sec:conclusion}
Adversarial machine learning research suffers from a lack of large scale, high quality datasets with realistic, physically plausible adversarial examples.
We overcome these limitations by using simulation to insert adversarial examples and generate synthetic data.
Our tools enables a new way for distributing datasets that are reproducible, configurable, extensible, and scalable.
Rather than downloading large, static archives, researchers can distribute a dataset in the form of configuration files and locally re-create the dataset by running our tools and CARLA.
They can also easily extend the dataset with new configurations based on their specific needs and dynamically insert adversarial examples in a realistic manner.
In future work, we would like to conduct large scale studies on the efficacy of attacks from various sources such as those shown in \cref{fig:ZoomedPatches}.
We believe this tool will enable more realistic evaluation of adversarial examples.

\ifdefined\isaccepted
\section*{Acknowledgements}
The work contained herein is developed under the DARPA Guaranteeing AI Robustness against Deception (GARD) program. The views, opinions and/or findings contained in this report are those of The MITRE Corporation and should not be construed as an official government position, policy, or decision, unless designated by other documentation. 

Approved for Public Release by MITRE; Distribution Unlimited. Public Release Case Number 22-2197.
This technical data deliverable was developed using contract funds under MITRE Basic Contract No. W56KGU-18-D-0004.

©2022 The MITRE Corporation. All rights reserved.
\fi

\bibliography{main}

\begin{thebibliography}{25}
\providecommand{\natexlab}[1]{#1}
\providecommand{\url}[1]{\texttt{#1}}
\expandafter\ifx\csname urlstyle\endcsname\relax
  \providecommand{\doi}[1]{doi: #1}\else
  \providecommand{\doi}{doi: \begingroup \urlstyle{rm}\Url}\fi

\bibitem[Braunegg et~al.(2020)Braunegg, Chakraborty, Krumdick, Lape, Leary,
  Manville, Merkhofer, Strickhart, and Walmer]{braunegg2020apricot}
Braunegg, A., Chakraborty, A., Krumdick, M., Lape, N., Leary, S., Manville, K.,
  Merkhofer, E., Strickhart, L., and Walmer, M.
\newblock {APRICOT}: A dataset of physical adversarial attacks on object
  detection.
\newblock In \emph{European Conference on Computer Vision}, pp.\  35--50.
  Springer, 2020.

\bibitem[Brown et~al.(2017)Brown, Mane, Roy, Abadi, and Gilmer]{46561}
Brown, T., Mane, D., Roy, A., Abadi, M., and Gilmer, J.
\newblock Adversarial patch.
\newblock 2017.
\newblock URL \url{https://arxiv.org/pdf/1712.09665.pdf}.

\bibitem[Carlini \& Wagner(2017)Carlini and Wagner]{Carlini2017}
Carlini, N. and Wagner, D.
\newblock Towards evaluating the robustness of neural networks.
\newblock In \emph{2017 IEEE Symposium on Security and Privacy (SP)}, pp.\
  39--57, 2017.
\newblock \doi{10.1109/SP.2017.49}.

\bibitem[Chen et~al.(2018)Chen, Cornelius, Martin, and
  Chau]{chen2018shapeshifter}
Chen, S.-T., Cornelius, C., Martin, J., and Chau, D. H.~P.
\newblock Shapeshifter: Robust physical adversarial attack on {Faster R-CNN}
  object detector.
\newblock In \emph{Joint European Conference on Machine Learning and Knowledge
  Discovery in Databases}, pp.\  52--68. Springer, 2018.

\bibitem[Cordts et~al.(2016)Cordts, Omran, Ramos, Rehfeld, Enzweiler, Benenson,
  Franke, Roth, and Schiele]{Cordts2016Cityscapes}
Cordts, M., Omran, M., Ramos, S., Rehfeld, T., Enzweiler, M., Benenson, R.,
  Franke, U., Roth, S., and Schiele, B.
\newblock The cityscapes dataset for semantic urban scene understanding.
\newblock In \emph{Proc. of the IEEE Conference on Computer Vision and Pattern
  Recognition (CVPR)}, 2016.

\bibitem[Cornelius et~al.(2019)Cornelius, Chen, Martin, and
  Chau]{Cornelius2019TalkPT}
Cornelius, C., Chen, S.-T., Martin, J., and Chau, D.~H.
\newblock Talk proposal: Towards the realistic evaluation of evasion attacks
  using {CARLA}.
\newblock \emph{ArXiv}, abs/1904.12622, 2019.

\bibitem[Crall(2020)]{kitware2020kwcoco}
Crall, J.
\newblock Kitware coco, 2020.
\newblock URL \url{https://gitlab.kitware.com/computer-vision/kwcoco}.

\bibitem[Dosovitskiy et~al.(2017)Dosovitskiy, Ros, Codevilla, Lopez, and
  Koltun]{Dosovitskiy17}
Dosovitskiy, A., Ros, G., Codevilla, F., Lopez, A., and Koltun, V.
\newblock {CARLA}: {An} open urban driving simulator.
\newblock In \emph{Proceedings of the 1st Annual Conference on Robot Learning},
  pp.\  1--16, 2017.

\bibitem[Dwibedi et~al.(2017)Dwibedi, Misra, and Hebert]{dwibedi2017cut}
Dwibedi, D., Misra, I., and Hebert, M.
\newblock Cut, paste and learn: Surprisingly easy synthesis for instance
  detection.
\newblock In \emph{Proceedings of the IEEE international conference on computer
  vision}, pp.\  1301--1310, 2017.

\bibitem[Fabbri et~al.(2021)Fabbri, Bras{\'o}, Maugeri, Cetintas, Gasparini,
  O{\v{s}}ep, Calderara, Leal-Taix{\'e}, and Cucchiara]{fabbri2021motsynth}
Fabbri, M., Bras{\'o}, G., Maugeri, G., Cetintas, O., Gasparini, R.,
  O{\v{s}}ep, A., Calderara, S., Leal-Taix{\'e}, L., and Cucchiara, R.
\newblock {MOTS}ynth: How can synthetic data help pedestrian detection and
  tracking?
\newblock In \emph{Proceedings of the IEEE/CVF International Conference on
  Computer Vision}, pp.\  10849--10859, 2021.

\bibitem[Feng et~al.(2022)Feng, Jha, and Prakash]{feng2022constraining}
Feng, R., Jha, S., and Prakash, A.
\newblock Constraining the attack space of machine learning models with
  distribution clamping preprocessing.
\newblock \emph{arXiv preprint arXiv:2205.08989}, 2022.

\bibitem[Geiger et~al.(2012)Geiger, Lenz, and Urtasun]{geiger2012we}
Geiger, A., Lenz, P., and Urtasun, R.
\newblock Are we ready for autonomous driving? the kitti vision benchmark
  suite.
\newblock In \emph{Computer Vision and Pattern Recognition (CVPR), 2012 IEEE
  Conference on}, pp.\  3354--3361. IEEE, 2012.
\newblock URL \url{https://ieeexplore.ieee.org/abstract/document/6248074}.

\bibitem[Goodfellow et~al.(2015)Goodfellow, Shlens, and
  Szegedy]{Goodfellow2015}
Goodfellow, I., Shlens, J., and Szegedy, C.
\newblock Explaining and harnessing adversarial examples.
\newblock In \emph{International Conference on Learning Representations}, 2015.
\newblock URL \url{http://arxiv.org/abs/1412.6572}.

\bibitem[Hendrycks \& Dietterich(2019)Hendrycks and
  Dietterich]{hendrycks2019benchmarking}
Hendrycks, D. and Dietterich, T.
\newblock Benchmarking neural network robustness to common corruptions and
  perturbations.
\newblock \emph{arXiv preprint arXiv:1903.12261}, 2019.

\bibitem[Johnson-Roberson et~al.(2017)Johnson-Roberson, Barto, Mehta, Sridhar,
  Rosaen, and Vasudevan]{Johnson-Roberson2017}
Johnson-Roberson, M., Barto, C., Mehta, R., Sridhar, S.~N., Rosaen, K., and
  Vasudevan, R.
\newblock Driving in the matrix: Can virtual worlds replace human-generated
  annotations for real world tasks?
\newblock In \emph{2017 IEEE International Conference on Robotics and
  Automation (ICRA)}, pp.\  746--753, 2017.
\newblock \doi{10.1109/ICRA.2017.7989092}.

\bibitem[Koh et~al.(2021)Koh, Sagawa, Marklund, Xie, Zhang, Balsubramani, Hu,
  Yasunaga, Phillips, Gao, et~al.]{koh2021wilds}
Koh, P.~W., Sagawa, S., Marklund, H., Xie, S.~M., Zhang, M., Balsubramani, A.,
  Hu, W., Yasunaga, M., Phillips, R.~L., Gao, I., et~al.
\newblock Wilds: A benchmark of in-the-wild distribution shifts.
\newblock In \emph{International Conference on Machine Learning}, pp.\
  5637--5664. PMLR, 2021.

\bibitem[Lu et~al.(2017)Lu, Sibai, Fabry, and Forsyth]{Lu2017NONT}
Lu, J., Sibai, H., Fabry, E., and Forsyth, D.~A.
\newblock No need to worry about adversarial examples in object detection in
  autonomous vehicles.
\newblock \emph{ArXiv}, abs/1707.03501, 2017.

\bibitem[Madry et~al.(2017)Madry, Makelov, Schmidt, Tsipras, and
  Vladu]{madry2017towards}
Madry, A., Makelov, A., Schmidt, L., Tsipras, D., and Vladu, A.
\newblock Towards deep learning models resistant to adversarial attacks.
\newblock \emph{arXiv preprint arXiv:1706.06083}, 2017.

\bibitem[Papernot et~al.(2016)Papernot, McDaniel, Jha, Fredrikson, Celik, and
  Swami]{Papernot2016}
Papernot, N., McDaniel, P., Jha, S., Fredrikson, M., Celik, Z.~B., and Swami,
  A.
\newblock The limitations of deep learning in adversarial settings.
\newblock In \emph{2016 IEEE European Symposium on Security and Privacy (EuroS
  P)}, pp.\  372--387, 2016.
\newblock \doi{10.1109/EuroSP.2016.36}.

\bibitem[Pintor et~al.(2022)Pintor, Angioni, Sotgiu, Demetrio, Demontis,
  Biggio, and Roli]{Pintor2022ImageNetPatchAD}
Pintor, M., Angioni, D., Sotgiu, A., Demetrio, L., Demontis, A., Biggio, B.,
  and Roli, F.
\newblock Imagenet-{P}atch: A dataset for benchmarking machine learning
  robustness against adversarial patches.
\newblock \emph{ArXiv}, abs/2203.04412, 2022.

\bibitem[Richter et~al.(2016)Richter, Vineet, Roth, and
  Koltun]{Richter2016PlayingFD}
Richter, S.~R., Vineet, V., Roth, S., and Koltun, V.
\newblock Playing for data: Ground truth from computer games.
\newblock \emph{ArXiv}, abs/1608.02192, 2016.

\bibitem[Sitawarin et~al.(2018)Sitawarin, Bhagoji, Mosenia, Mittal, and
  Chiang]{sitawarin2018rogue}
Sitawarin, C., Bhagoji, A.~N., Mosenia, A., Mittal, P., and Chiang, M.
\newblock Rogue signs: Deceiving traffic sign recognition with malicious ads
  and logos.
\newblock \emph{arXiv preprint arXiv:1801.02780}, 2018.

\bibitem[Szegedy et~al.(2018)Szegedy, Zaremba, Sutskever, Bruna, Erhan,
  Goodfellow, and Fergus]{Szegedy2014}
Szegedy, C., Zaremba, W., Sutskever, I., Bruna, J., Erhan, D., Goodfellow, I.,
  and Fergus, R.
\newblock Intriguing properties of neural networks.
\newblock In \emph{Joint European Conference on Machine Learning and Knowledge
  Discovery in Databases}, pp.\  52--68. Springer, 2018.

\bibitem[Threet et~al.(2021)Threet, Busho, Harguess, Jutras, Lape, Leary,
  Manville, Tan, and Ward]{9762099}
Threet, M., Busho, C., Harguess, J., Jutras, M., Lape, N., Leary, S., Manville,
  K., Tan, M., and Ward, C.
\newblock Physical adversarial attacks in simulated environments.
\newblock In \emph{2021 IEEE Applied Imagery Pattern Recognition Workshop
  (AIPR)}, pp.\  1--5, 2021.
\newblock \doi{10.1109/AIPR52630.2021.9762099}.

\bibitem[Voigtlaender et~al.(2019)Voigtlaender, Krause, Osep, Luiten, Sekar,
  Geiger, and Leibe]{voigtlaender2019mots}
Voigtlaender, P., Krause, M., Osep, A., Luiten, J., Sekar, B. B.~G., Geiger,
  A., and Leibe, B.
\newblock Mots: Multi-object tracking and segmentation.
\newblock In \emph{Proceedings of the IEEE/CVF Conference on Computer Vision
  and Pattern Recognition}, pp.\  7942--7951, 2019.

\end{thebibliography}
\bibliographystyle{icml2022}

\newpage
\appendix

\onecolumn

\section{Appendix}
\label{sec:appendix}
\begin{figure}[ht]
    \subfigure[Sunny]{\label{fig:weather_sunny}\includegraphics[width=0.32\textwidth]{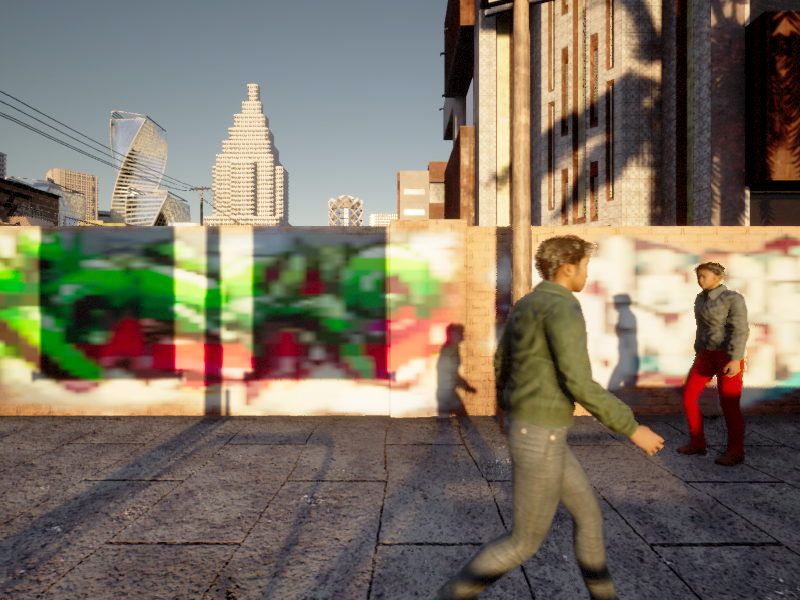}}
    \hfill
    \subfigure[Rainy]{\label{fig:weather_rainy}\includegraphics[width=0.32\textwidth]{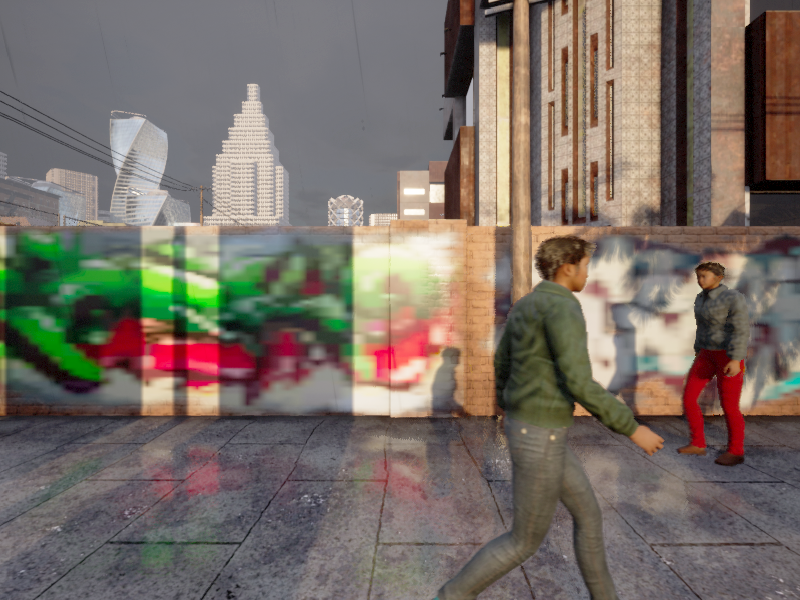}}
    \hfill
    \subfigure[Foggy]{\label{fig:weather_foggy}\includegraphics[width=0.32\textwidth]{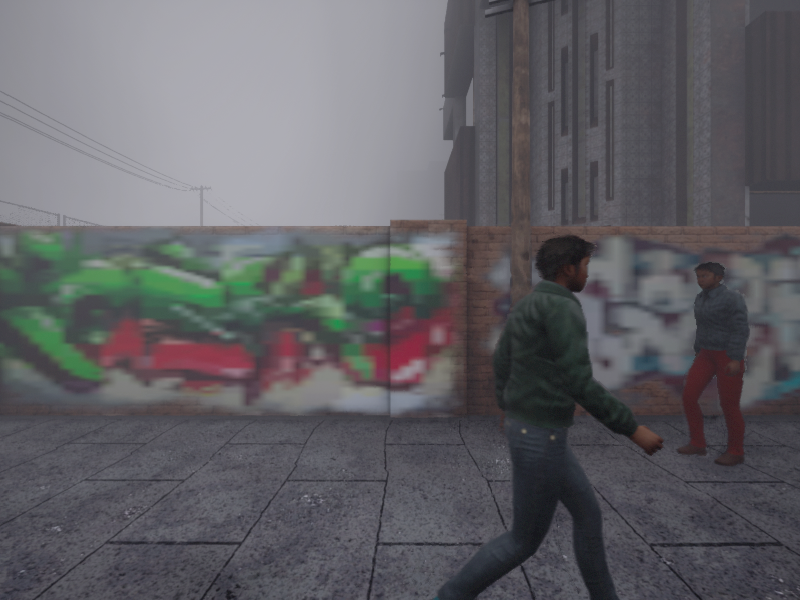}}
    \caption{Our tool can repeatably collect data over a weather distribution shift. We collected frames from the same scenario for different weather conditions: \subref{fig:weather_sunny}~sunny, \subref{fig:weather_rainy}~rainy, and \subref{fig:weather_foggy}~foggy.}
  \label{fig:weather_change}
\end{figure}

The synthetic data collection tool described in \cref{sec:tool} consists of two major components, a data saver and an annotator.
The data saver takes a configuration file as input to configure the CARLA simulation scenario.
This configuration includes CARLA server configurations, weather settings, and configurations for spawning actors (i.e., vehicles and pedestrians) and sensors for data capture. 
\Cref{lst:config_yaml} shows an example of such a configuration that captures $300$ frames from three sensors (RGB camera, depth camera, and instance segmentation sensor) with two pedestrians.
These sensors are statically placed to capture the two pedetrians as they walk past these sensors.
The two pedestrians that walk past each other are specified by their the start positions, destination positions, and walking speeds.
Finally, the configuration sets a seed to enable repeated generations of a particular scenario.
\Cref{fig:weather_change} shows example frames captured from three different captures under different weather conditions using configurations shown in \Cref{lst:config_yaml} and \Cref{lst:weather_config}.
Our tool enables reproducible capture under a variety of conditions like weather and lighting.

\definecolor{lightgray}{rgb}{0.9,0.9,0.9}
\begin{lstlisting}[language=yaml,
                  numbers=left,
                  frame=lines,
                  framesep=1em,
                  backgroundcolor=\color{lightgray},
                  framexleftmargin=0.05\textwidth,framexrightmargin=0.05\textwidth,
                  xleftmargin=.06\textwidth,xrightmargin=.06\textwidth,
                  basicstyle=\footnotesize\ttfamily,
                  framexleftmargin=1cm,
                  framexrightmargin=1cm,
                  caption={An example \texttt{config.yaml} file that generates 2 pedestrians with 3 static sensors (RGB, depth and instance segmentation).},
                  label=lst:config_yaml]
carla:
  host: "127.0.0.1"
  port: 2000
  timeout: 5.0
  sync:
    fps: 30
    timeout: 2.0
  seed: 30
  townmap: "Town10HD"
  traffic_manager_port: 8000
  retry: 10

output_dir: "_out"
max_frames: 300

# Sunny weather
weather:
  cloudiness: 0.0
  precipitation: 0.0
  precipitation_deposits: 0.0
  wind_intensity: 0.0
  sun_azimuth_angle: 0.0
  sun_altitude_angle: 10.0
  fog_density: 0.0
  fog_distance: 0.0
  wetness: 0.0

spawn_actors:
  # Spawn Pedestrian 1
  - blueprint:
      name: "walker.pedestrian.*"
      attr: {role_name: "hero1", is_invincible: "false"}
      speed: 1.4  # Between 1 and 2 m/s (default is 1.4 m/s).
    transform:
      location: {x: -91, y: 170, z: 0.6}
      rotation: {yaw: -90.0}
    destination_transform:
      location: {x: -91, y: 150, z: 0.6}
  # Spawn statically placed sensors - RGB, Depth and Instance Segmentation
  - blueprint:
      name: "sensor.camera.rgb"
      attr: {"image_size_x": "800", "image_size_y": "600"}
    transform:
      location: {x: -95, y: 160, z: 1.6}
      rotation: {yaw: 0.0}
  - blueprint:
      name: "sensor.camera.depth"
      attr: {"image_size_x": "800", "image_size_y": "600"}
    transform:
      location: {x: -95, y: 160, z: 1.6}
      rotation: {yaw: 0.0}
  - blueprint:
      name: "sensor.camera.instance_segmentation"
      attr: {"image_size_x": "800", "image_size_y": "600"}
    transform:
      location: {x: -95, y: 160, z: 1.6}
      rotation: {yaw: 0.0}
  # Spawn Pedestrian 2
  - blueprint:
      name: "walker.pedestrian.*"
      attr: {role_name: "hero2", is_invincible: "false"}
      speed: 2.0
    transform:
      location: {x: -91, y: 150, z: 0.6}
      rotation: {yaw: 90.0}
    destination_transform:
      location: {x: -91, y: 170, z: 0.6}
\end{lstlisting}

\begin{lstlisting}[language=yaml,
                  numbers=left,
                  frame=lines,
                  framesep=1em,
                  backgroundcolor=\color{lightgray},
                  framexleftmargin=0.05\textwidth,framexrightmargin=0.05\textwidth,
                  xleftmargin=.06\textwidth,xrightmargin=.06\textwidth,
                  basicstyle=\footnotesize\ttfamily,
                  framexleftmargin=1cm,
                  framexrightmargin=1cm,
                  caption={An example to show 2 different weather settings used to create the examples shown in~\cref{fig:weather_change}},
                  label=lst:weather_config]
# Rainy weather
weather:
  cloudiness: 60.0
  precipitation: 60.0
  precipitation_deposits: 60.0
  wind_intensity: 60.0
  sun_azimuth_angle: -1.0
  sun_altitude_angle: 15.0
  fog_density: 3.0
  fog_distance: 0.75
  wetness: 0.0
  
# Foggy weather
weather:
  cloudiness: 0.0
  precipitation: 0.0
  precipitation_deposits: 0.0
  wind_intensity: 0.0
  sun_azimuth_angle: 0.0
  sun_altitude_angle: 10.0
  fog_density: 100.0
  fog_distance: 1.0
  wetness: 0.0
\end{lstlisting}

\end{document}